\documentclass[10pt,twocolumn,letterpaper]{article}

\usepackage{iccv}

\usepackage{booktabs}
\usepackage{multirow}

\input CSMacrosV2.tex

\usepackage{hyperref}
\hypersetup{
    colorlinks=true,
    breaklinks=true,
    urlcolor=blue,
    linkcolor=blue,
    bookmarksopen=false,
    filecolor=black,
    citecolor=blue,
    linkbordercolor=blue
}

\iccvfinalcopy %

\def\iccvPaperID{6529} %

\begin{document}

\title{Channel-wise  Knowledge Distillation for Dense Prediction\thanks{C. Shu and Y. Liu contributed equally. Accepted to Proc.\ Int.\ Conf.\ Computer Vision 2021.
\protect\\
\hspace*{1.35em}$^1$Shanghai Em-Data Technology Co.\protect\\
\hspace*{1.35em}$^2$The University of Adelaide, Australia\protect\\
\hspace*{1.35em}$^3$Monash University, Australia\protect\\
\hspace*{1.35em}$^4$Baidu Inc.
}
}

\author{Changyong Shu$^{1,4}$, ~~~
Yifan Liu$^2$, ~~~~
Jianfei Gao$^1$, ~~~~
Zheng Yan$^1$
, ~~~~
Chunhua Shen$^3$
}

\makeatletter
\let\@oldmaketitle\@maketitle%
\renewcommand{\@maketitle}{\@oldmaketitle%
 \centering
    \includegraphics[trim=0 .25cm 0 0,clip,width=.869\linewidth]{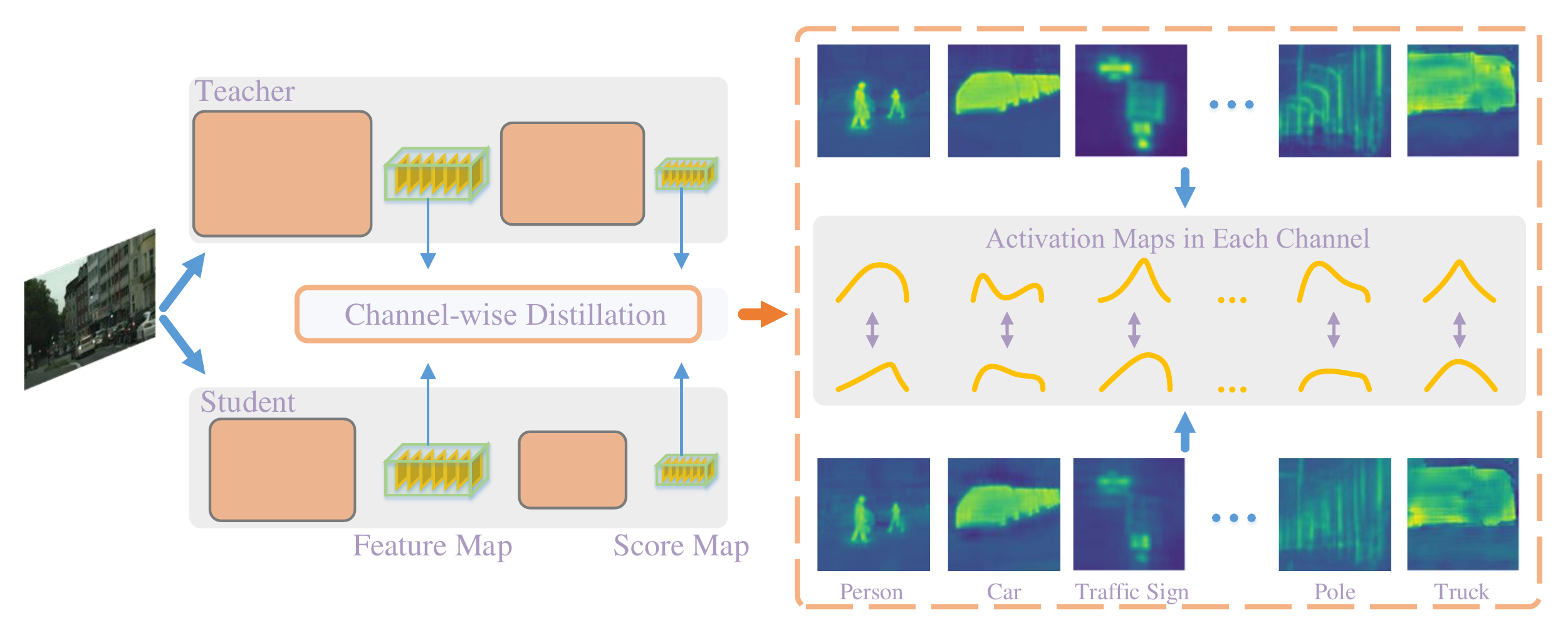}
    \captionof{figure}{
\textbf{The
	    overall
	    architecture of our proposed method}.
	    The plot
	on the left %
	shows
	our teacher-student strategy, where the feature map and the logits map can be used for channel-wsie %
	knowledge
	distillation. The
	plot
	on the right
	shows an intuitive illustration:
	activated regions correspond to scene categories.
    }
    \label{fig:overview}
    \bigskip}                   %
\makeatother

\maketitle
\ificcvfinal\thispagestyle{empty}\fi

\begin{abstract}
    Knowledge distillation (KD) has been proven %
    a simple and effective tool for training compact dense prediction models. Lightweight student networks are trained by extra %
    supervision
    transferred from large teacher networks. Most %
    previous KD variants for dense prediction tasks align the activation maps from the student and teacher network in the spatial domain, typically by normalizing the activation values on each spatial location and minimizing point-wise and/or pair-wise discrepancy.
    Different from the previous methods, here we propose to normalize the activation map %
    of
    each channel to obtain a soft %
    probability map.
    By simply minimizing the Kullback–Leibler (KL) divergence between the channel-wise  %
    probability map
    of the two networks, the distillation process pays more attention to the most salient regions of each channel, which are %
    valuable
    for dense prediction tasks.

    We conduct experiments on a few %
    dense prediction tasks, including semantic segmentation and object detection.
    Experiments demonstrate that our %
    proposed method
    outperforms state-of-the-art distillation methods considerably, and can require less computational cost during training. In particular, we improve the RetinaNet detector %
    (ResNet50 backbone) by $3.4\%$ in %
    mAP on the COCO dataset, and PSPNet %
    (ResNet18 backbone) by $5.81\%$ in %
    mIoU on the Cityscapes dataset.
    Code %
    is available at:

    \url{https://git.io/Distiller}

\end{abstract}

\section{Introduction}

\begin{figure}[htb]
\centering
      \includegraphics[width=1.0\linewidth]{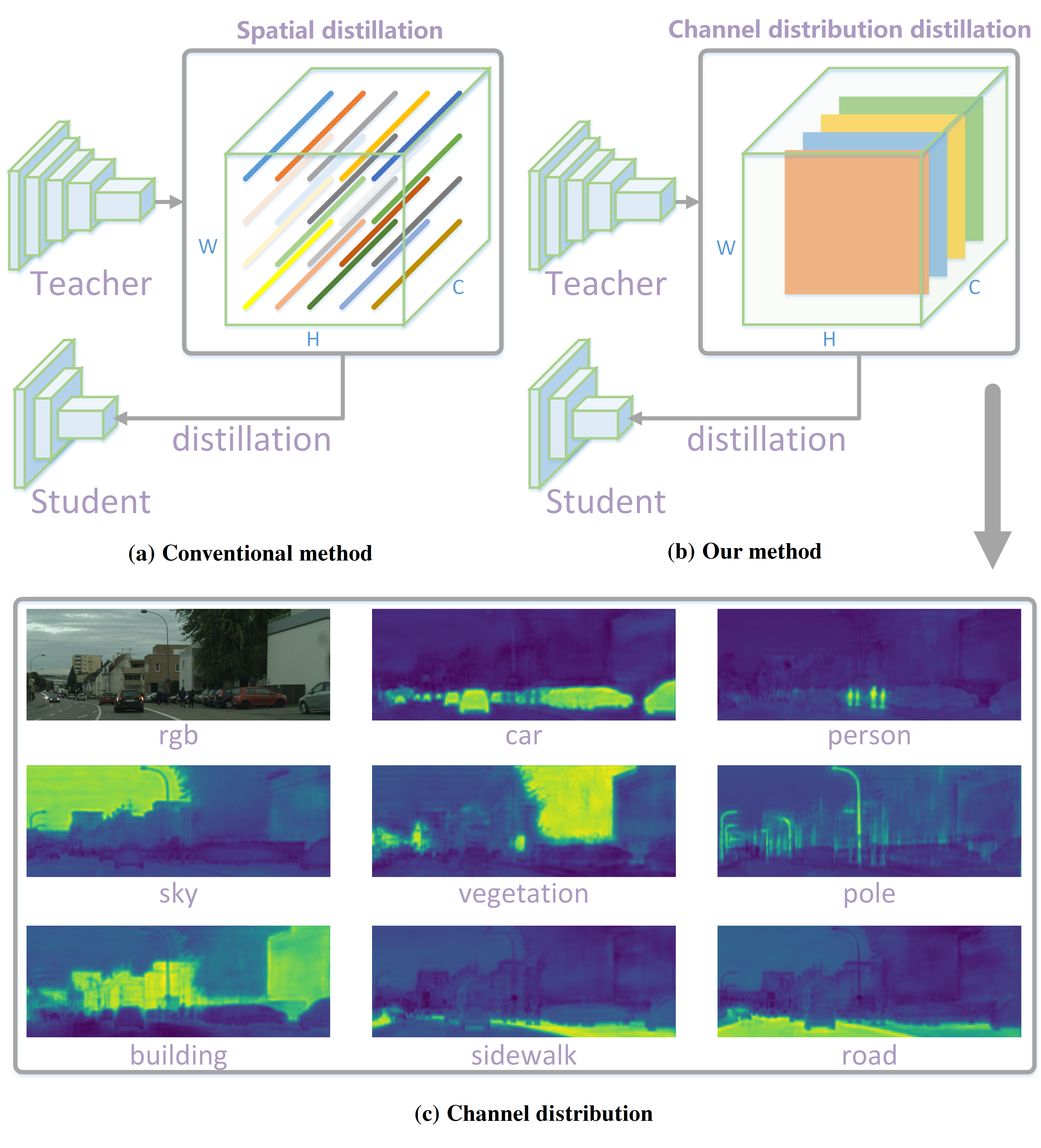}
   \caption{\textbf{Spatial knowledge distillation} (top-left) works by aligning feature maps  in the spatial domain.
   \textbf{Our channel distribution distillation} (top-right) instead aligns each channel of the student's feature maps to
   that of the teacher network by minimizing the KL divergence.
   The bottom plot shows that the
   activation values of each channel tend to
   encode
   saliency
   of scene categories.
   }
   \label{fig:intro}
\end{figure}

Dense prediction tasks are a group of fundamental tasks in computer vision, including semantic segmentation~\cite{zhao17pspnet,Chen17deeplabv3} and object detection~\cite{lin2017retinanet,zhi2019fcos}. These tasks require learning strong feature representations for complex scene understanding at the pixel level.
Thus, %
state-of-the-art models usually need high computational costs, making them %
cumbersome
to be deployed to %
mobile devices. %
As a result, compact networks designed for dense prediction tasks have drawn much attention. Moreover, effectively
training lightweight networks has been studied in previous works %
using
knowledge distillation (KD). A compact network is trained
with the supervision of a large teacher network, and can achieve
better performance. Pioneering works~\cite{hinton14distilling, adriana15fitnets} are proposed and well studied, mostly for image classification tasks.

Dense prediction tasks are per-pixel prediction problems, which are more %
challenging than image-level classification. Previous research~\cite{liu19structuredDense,li17mimicking} found that directly transferring the KD methods~\cite{hinton14distilling,adriana15fitnets} in classification to semantic segmentation may not lead to satisfactory results. \textit{Strictly aligning the point-wise classification scores or the feature maps between the teacher and student network may enforce overly strict constraints and lead to sub-optimal solutions.}

Recent works \cite{liu19structuredDense,liu2019structuredseg,hou20inter} pay attention to enforce the correlations among different spatial locations.  As shown in Figure \ref{fig:intro}(a), the activation values\footnote{The activation values in this work include the final logits and the inner feature maps.} on each spatial location are normalized. Then, some tasks specific relationships are conducted by aggregating a sub-set of different spatial locations, such as pair-wise relations~\cite{liu19structuredDense, xie18improving} and inter-class relations~\cite{hou20inter}. Such methods may
work better than the point-wise alignment in capturing spatial structure information and improve the performance of the student network. However, every spatial location in the activation map contributes equally to the knowledge transferring, which may bring redundant information from the teacher network.

In this work, we propose a novel channel-wise
knowledge
distillation  by normalizing the activation maps in each channel for dense prediction tasks, as shown in Figure \ref{fig:intro}(b). Then,
we minimize the
asymmetry Kullback–Leibler (KL) divergence
of the normalized channel %
activation maps---which is converted into a  distribution for each channel---%
between
the teacher and the student networks. We show an example of the channel-wise  distribution
in Figure \ref{fig:intro}(c). The activations of each channel tend to encode saliency %
of scene categories.
For
each channel, the student %
network is guided to
pay more attention to mimic the regions with %
significant
activation values, %
leading  to a more accurate localization in dense prediction tasks. For example, in object detection, the student network %
pays more attention to learn the activations of the foreground objects.

Some recent works %
exploit
knowledge contained in channels. Channel distillation \cite{zhou20channel} proposes to transfer the activation in each channel into one %
aggregated scalar, which may be helpful for image-level classification, %
but the spatial aggregation loses all spatial information and thus is not suitable for dense prediction. Other works, such as MGD \cite{yue2020matching}, Channel exchanging \cite{wang2020deep} and CSC~\cite{park2020knowledge}
show
the importance of  channel-wise  information.
MGD matches the teacher channels with
students' and solves it as an assignment problem.
 Channel exchanging \cite{wang2020deep} uses a
  fusion module to dynamically exchange channels
between sub-networks of various modalities.

 We
 show that the simple normalizing operations for each channel can improve the baseline spatial distillation by a large margin.
 The proposed channel-wise
 distillation is simple and easy to %
 apply
 to %
 various
 tasks and network structures.
We summarize our main contributions as follows.
\begin{itemize}
\itemsep -0.1cm

    \item Unlike those existing spatial distillation approaches, we propose a novel channel-wise
    distillation paradigm
    for dense prediction tasks. Our method is simple yet effective.

    \item The proposed channel-wise
    distillation significantly outperforms
    state-of-the-art KD  methods for semantic segmentation and object detection.

    \item  We show consistent improvements on four
    benchmark datasets  with
    various
    network structures on semantic segmentation and object detection tasks,
    demonstrating
    that our method is general. Given its  simplicity and effectiveness,
    we believe that our method
    can serve as
    a strong baseline KD method for dense prediction tasks.
\end{itemize}

\section{Related Work}
Most
works on knowledge distillation focus on classification tasks \cite{%
jie2020rolekd,guan2020dfas,hinton14distilling,park2019relational,Xie2020sns,
yuan2020revisiting,zhang2019distill}.
Our work here aims to study efficient and effective distillation methods for dense prediction,
beyond naively
 applying pixel-wise distillation as done in classification.

\noindent\textbf{Knowledge distillation for semantic segmentation.}
In \cite{xie18improving},
a local similarity map is constructed to minimize the discrepancy of segmented boundary information between the teacher and student network, where the Euclidean distance between the center pixel and the 8-neighborhood pixels is used as knowledge for transferring. Liu \etal
\cite{liu2019structuredseg,liu19structuredDense}
propose two
approaches
to capture the structured information among pixels, including pair-wise similarity between pixels and holistic correlations captured by a discriminator.
The work in \cite{wang20intraclass} %
focuses on the intra-class feature variation among the pixels with the same label, where the set of cosine distance between each pixel’s feature and its corresponding class-wise prototype is constructed to transfer the structural knowledge.
He \etal \cite{he19knowledge} use a feature adaptor is employed to mitigate the feature mismatching between the teacher and student networks.

\begin{table*}[htb]
	\renewcommand\arraystretch{1.1}

	\footnotesize

	\centering

	\begin{tabular}{p{4.0cm}|c|c|c}
    	\hline
		\multirow{2}*{Loss}  &\multirow{2}*{\textcolor{white}{*******}$\varphi(u, v)$\textcolor{white}{*******}} &\multicolumn{2}{c}{$\phi(x)$} \\
		\cline{3-4}
		&  &\textcolor{white}{*******}{Formulation}\textcolor{white}{*******} &\textcolor{white}{*****}Dimensionality\textcolor{white}{*****} \\

		          	\hline
         \multicolumn{4}{c}{%
         Point-wise alignment}\\
             	\hline
		\multirow{1}*{Attention transfer \cite{sergey17paying}}
		 &$L_1$ or $L_2$
		&
		${\sum_{c=1}^{C}\|x_{ic}\|^{p}}$
		&\multicolumn{1}{c}{${1 \times W \times H}$} \\
		\multirow{1}*{Pixelwise \cite{liu19structuredDense,chen20fasterseg,liu2019structuredseg,wang20intraclass}}        &KL            &${\textup{softmax}(x_{i}/\tau)}$                     &\multicolumn{1}{c}{${C \times W \times H}$} \\
		    	          	\hline
         \multicolumn{4}{c}{%
         Pairwise or higher order alignment }\\
             	\hline
		\multirow{1}*{Local similarity \cite{xie18improving}}
		       &$L_1$ or $L_2$                 &${\sum_{j\in N(i)}\|x_{j}-x_{i}\|}$                  &\multicolumn{1}{c}{${1 \times W \times H}$} \\
		\multirow{1}*{Pairwise affinity \cite{liu19structuredDense,he19knowledge,liu2019structuredseg}}
		 &$L_2$            &$\frac{x^T_{i}x_j}{\|x_i\|_2 \cdot \|x_j\|_2} $                 &\multicolumn{1}{c}{${1 \times W \times H}$} \\
		\multirow{1}*{IFVD \cite{wang20intraclass}}   &$L_2$                       &${\cos(x_i, \sum_{j\in S_i}^{}x_j/|S_i|)}$ &\multicolumn{1}{c}{${1 \times W \times H}$} \\
		\multirow{1}*{Holistic \cite{liu19structuredDense,liu2019structuredseg,%
		wang20intraclass}}
		 &Wasserstein Distance                &${D(x_{i})}$                                         &\multicolumn{1}{c}{${1}$} \\
		    	\hline
	\end{tabular}
		\caption{Current spatial distillation methods. $i$ and $j$ indicate the pixel index.
		$D(\cdot )$ is a discriminator, and $N(i)$ indicates 8-neighborhood of pixel $i$.
		$S_i$ is the pixel set having the same label as pixel $i$
		and $|S_i|$ stands for the size of the set $S_i$.
		}
	\label{tab:pixel-table}
\end{table*}
\noindent\textbf{Knowledge distillation for object detection.}
Many methods find that it is important to distinguish the foreground and the background regions in the distillation %
for object detection.
MIMIC \cite{li17mimicking} forces the feature map inside the RPN of the student network to be similar to that of the teacher network via the
$L_2$ loss,  and finds that directly applying pixel-wise loss may harm the performance of object detection.  Wang \etal  \cite{Wang2019dod} propose to distill the fine-grained feature near object anchor locations. Zhang and Ma
\cite{Zhang2021iod} generate the mask with attention to distinguish the foreground and the background, achieving promising results.
Instead, we softly align the channel-wise activations to
distinguish the foreground and the background regions.

\noindent\textbf{Channel-wise knowledge.} Several recent  works~\cite{zhou20channel} also pay attention to the knowledge %
contained
in each channel. Zhou \etal calculate the mean of the activation in each channel and align a weighted difference for each channel in classification.
CSC~\cite{park2020knowledge} calculates the pair-wise relations among all spatial locations and all channels for transferring the knowledge.
Channel exchanging~\cite{wang2020deep} %
proposes that the information contained in each channel is general and can be shared across different modalities.

\section{Our Method}

 We first review relevant spatial knowledge distillation methods
 in the literature.
\subsection{Spatial Distillation}\label{arch-KSANC}
Existing KD methods often employ a point-wise alignment or
align
structured %
information among spatial locations, which can be formulated as:
\begin{equation}
	\ell (y, y^{S})+{\alpha}
	\cdot
	\varphi(\phi( {y}^{T}), \phi( {y}^{S})).
\end{equation}
Here
the task loss
$\ell (\cdot )$
is still applied with $ y $ being the ground-truth labels.
For example, the cross-entropy loss is usually employed in semantic segmentation.
By slightly abusing the notation, here $ y^S $ and $ y^T $ represent either the logits or inner activations of the student
and teacher network, respectively.
Here
$ \alpha $ is a hyper-parameter %
to balance the loss terms.
Subscripts $\cdot ^T $ and $ \cdot ^S$ denote teacher and student networks.
We list representative spatial distillation methods
in Table \ref{tab:pixel-table}.

A brief overview of these methods is as follows.
Attention Transfer (AT) \cite{sergey17paying} uses an attention mask to squeeze the feature maps into a single channel for distillation. The pixel-wise loss~\cite{hinton2015distilling} directly aligns the point-wise class probabilities. The local affinity~\cite{xie18improving} is computed by the distance between the center pixel and its $8$ neighborhood pixels. The pairwise affinity~\cite{liu19structuredDense,he19knowledge,liu2019structuredseg} is employed to transfer the similarity between pixel pairs. The similarity between each pixel’s feature and its corresponding class-wise prototype is computed to transfer the structural knowledge~\cite{wang20intraclass}. The holistic loss in \cite{liu19structuredDense,
liu2019structuredseg} uses the adversarial scheme to align the high-order relations between feature maps from the two networks. Note that, the last four terms consider the correlation among pixels. Existing KD methods as shown in Table \ref{tab:pixel-table} are all spatial distillation methods. All these methods consider the $N$ channel activation values of a spatial location as the feature vectors to operate on.

\subsection{Channel-wise
Distillation}\label{channel-wise-alignment}
To better
exploit
the knowledge in each channel, we propose to
\textit{softly} align activations of corresponding channels between
the teacher and student networks.
To do so, we first convert activations of a channel into
a probability distribution  such that we can measure the discrepancy using a probability distance metric such as the KL divergence.
As demonstrated  in
Figure \ref{fig:intro}(c), the activations of different channels tend to encode the saliency of scene categories of an input image. Besides, a well-trained teacher network
for semantic segmentation shows
activation maps of clear
category-specific masks for each channel---which is expected---%
as %
displayed
on the right part of Figure \ref{fig:overview}.
Here,
we propose a novel channel-wise
distillation paradigm to guide the student to %
learn %
the knowledge
from a well-trained teacher.

Let us denote the teacher and student networks %
as $T$ and $S$, and the activation maps from $T$ and $S$ are $y^{T}$ and $y^{S}$, respectively.  The channel-wise
distillation loss can be formulated as in a general form:
\begin{equation}
	\varphi(\phi({\it{y}^{T}}), \phi({\it{y}^{S}})) = \varphi(\phi({\it{y}^{T}_{c}}), \phi({\it{y}^{S}_{c}})).
\label{eq:cw}
\end{equation}
In our case,
$\phi(\cdot ) $ is used to %
convert
the activation values into
a probability distribution as below:
\begin{equation}
	\phi%
	{\left(y_{c}\right) } =  \frac{\textup{exp}{(\frac{y_{c,i}}{\mathcal{T}} )}} {\sum_{i = 1}^{W\cdot H} \textup{exp}{(\frac{y_{c,i}}{\mathcal{T}} )} } ,
	\label{eq:cw-phi}
\end{equation}
where $c = 1,2,..., C$ %
indexes the channel; and $ i $ indexes the spatial location of a channel.
$\mathcal{T}$ is a hyper-parameter (the temperature). The
probability becomes
softer if we use a larger $\mathcal{T}$, %
meaning that  we focus on a wider spatial region for  each channel. By applying the softmax normalization, we remove the influences of %
magnitude scales between the large networks and the compact networks.
This normalization is helpful in KD
as observed in
\cite{wang2020defense}. A $1\times1$ convolution layer %
is
employed to upsample  the number of channels for the student network if the number of channels %
mismatches
between the teacher and the student.
$\varphi (\cdot ) $ evaluates the discrepancy between the channel distribution from the teacher network and the student network. We use the KL divergence:
\begin{equation}
	\varphi%
	{\left(y^{T}, y^{S}\right) } = \frac{\mathcal{T}^2}{C}\sum_{c = 1}^{C}\sum_{i=1}^{W\cdot H}
	\phi (y^{T}_{c,i}) \cdot \log \Bigl[
	\frac{\phi(y^{T}_{c,i})}{\phi(y^{S}_{c,i})}
	\Bigr].
	\label{eq:cw2}
\end{equation}
The KL divergence is an asymmetric metric. From Equation \eqref{eq:cw2}, we can see that,
if $\phi (y^{T}_{c,i})$ is large,
$\phi(y^{S}_{c,i})$ should be as large as $\phi (y^{T}_{c,i})$ to minimize the KL divergence. Otherwise, if $\phi (y^{T}_{c,i})$ is very small, the KL divergence %
pays less attention to minimize the $\phi(y^{S}_{c,i})$. Thus, the student network tends to produce similar activation distribution in the foreground
saliency,
while the activations corresponding to the background region of the teacher network would have less impact on  the learning.
We hypothesize that this asymmetry property of KL benefits the KD learning for dense prediction tasks.

\section{Experiments}

In this section, we first describe the implementation details and the experiment settings. Then, we compare our channel-wise
distillation method with other state-of-the-art distillation methods and conduct ablation studies on semantic segmentation. Finally, we show consistent improvements in semantic segmentation and object detection with various benchmarks and student network structures.

\begin{table*}[htb]%
	\renewcommand\arraystretch{1.1}
	\footnotesize
	\centering

	\begin{tabular}{c|p{2.5cm}|c|c|c|c}
    \hline

		\multicolumn{2}{c|}{\multirow{2}{*}{Network}} &\multirow{2}{*}{Structural} &\multicolumn{1}{c|}{\multirow{2}{*}{\textcolor{white}{**}Complexity\textcolor{white}{**}}} &\multicolumn{2}{c}{Val mIoU (\%)} \\

		\cline{5-6}
      \multicolumn{2}{l|}{} &\multicolumn{1}{l|}{} & &\multicolumn{1}{c|}{Feature map} &\multicolumn{1}{c}{Logits map} \\

      \hline
      \multicolumn{2}{c|}{Teacher} &\multicolumn{1}{c|}{$-$} &\multicolumn{1}{c|}{$-$} &\multicolumn{1}{c|}{\textcolor{white}{*****}$78.56$\textcolor{white}{*****}} & \multicolumn{1}{c}{\textcolor{white}{*****}$78.56$\textcolor{white}{*****}}   \\
		\multicolumn{2}{c|}{Student} &\multicolumn{1}{c|}{$-$} &\multicolumn{1}{c|}{$-$} &\multicolumn{1}{c|}{$69.10$} & \multicolumn{1}{c}{$69.10$}   \\
      \hline
          \multirow{6}{*}{Spatial Distillation}

                                         &AT \cite{sergey17paying}                                                             &\multicolumn{1}{c|}{$\times$} &$h_{x}\cdot w_{x}\cdot (c_{x})^{p}$  &$72.37(+3.27)^{\circledast}$          &$72.32(+3.22)$   \\
                                         &PI \cite{chen20fasterseg,wang20intraclass,liu19structuredDense,liu2019structuredseg} &\multicolumn{1}{c|}{$\times$} &$h_{x}\cdot w_{x}\cdot c_{x}$        &$70.02(+0.92)^{\circledast}$          &$71.74(+2.64)$   \\
         \cline{2-6}
                                         &LOCAL \cite{xie18improving}                                                          &\multicolumn{1}{c|}{$\surd$}  &$8h_{x}\cdot w_{x}\cdot c_{x}$        &$69.81(+0.71)\textcolor{white}{^{\circledast}}$ &$69.75(+0.65)$   \\
                                         &PA \cite{liu19structuredDense,he19knowledge,liu2019structuredseg}                    &\multicolumn{1}{c|}{$\surd$}  &$(h_{x}\cdot w_{x})^{2}\cdot c_{x}$  &$71.23(+2.13)\textcolor{white}{^{\circledast}}$ &$71.41(+2.31)$   \\
                                         &IFVD \cite{wang20intraclass}                                                         &\multicolumn{1}{c|}{$\surd$}  &$h_{x}\cdot w_{x}\cdot c_{x}\cdot n$       &$71.35(+2.25)\textcolor{white}{^{\circledast}}$ &$70.66(+1.56)$   \\
                                         &HO \cite{liu19structuredDense,liu2019structuredseg,%
         wang20intraclass}       &\multicolumn{1}{c|}{$\surd$}  &$\mathcal O(D)$          &$-^{\circledast}$                               &$72.13(+3.03)$   \\
         \cline{1-6}
         \multirow{1}{*}{Channel Distillation}  &CD (Ours)     &\multicolumn{1}{c|}{$\surd$}  &$h_{x}\cdot w_{x}\cdot c_{x}$                                &$\textbf{74.27}(+5.17)^{\circledast}$           &$\textbf{74.87}(+5.77)$ \\
      \hline
	\end{tabular}
	\caption{Comparison between computation complexity and performance on the validation set among various distillation methods. The mIoU is calculated on the Cityscapes validation set with PSPNet-R101 as the teacher network and PSPNet-R18 as the student network.
	The complexity depends on the shape ($h_x \times w_x\times c_x$) of the input. $\mathcal O(D)$ denotes the discriminator complexity. The superscript $\circledast$ means that additional channel alignment convolution is needed. All the results are the mean of three runs.
	}
	\label{tab:individual-competition-table}
\end{table*}
\subsection{Experimental Settings}
\label{sec:imp}
\noindent\textbf{Datasets.} Three %
public
semantic segmentation benchmarks, %
namely,
Cityscapes~\cite{marius16cityscape}, ADE20K \cite{zhou17ade20k} and Pascal VOC~\cite{mark10voc} are %
used here.
We also apply the proposed distillation method to object detection on MS-COCO 2017~\cite{lin2014microsoft}, which is a large-scale dataset that contains over 120k images %
of
80 categories.

The Cityscapes dataset is used for semantic urban scene understanding. It contains 5,000 finely annotated images with 2,975/500/1,525 images for training/validation/testing respectively, where 30 common classes are provided and 19 classes are used for evaluation and testing. The size of each image is $2048 \times 1024$ pixels. %
They are %
gathered from 50 different cities. The coarsely labeled data is not used in our experiments.

The Pascal VOC dataset contains 1,464/1,449/1,456 images for training/validation/testing. It contains 20 foreground object classes and an extra background class. In addition, the dataset is augmented by extra coarse labeling, which has 10,582 images for training. The training split is used for training, and the final performance is measured on the validation set across 21 classes.

The ADE20K dataset covers 150 classes of diverse scenes.
It contains 20K/2K/3K images for training, validation, and testing. In our experiments, we report the segmentation accuracy on the validation set.

\noindent\textbf{Evaluation metrics.} To evaluate the performance and efficiency of our proposed channel distribution distillation method on semantic segmentation, following the previous work~\cite{hou20inter,liu2019structuredseg}, we test each strategy via the mean Intersection-over-Union (mIoU)
in all experiments under a single-scale setting.
The floating-point operations per second (FLOPs) are calculated with a fixed input size of 512$\times$1024 pixels. Besides, the mean class Accuracy (mAcc) is listed for Pascal VOC and ADE20K. To evaluate the performance on object detection, we report the mean Average Precision (mAP), the inference speed (FPS), and the model size (parameters) following the work in \cite{Zhang2021iod}.

\noindent\textbf{Implementation details.} For semantic segmentation, the teacher network is PSPNet with ResNet101 (PSPNet-R101) as the backbone for all experiments. We employ several different architectures, including PSPNet~\cite{zhao17pspnet}, Deeplab~\cite{zhang18deep} with the backbones of ResNet18, and MobileNetV2 as student networks to verify the effectiveness of
our method.

In the ablation study, we analyze the effectiveness of our method based on PSPNet with the ResNet18 backbone (PSPNet-R18). Unless otherwise indicated, each training image for the student network is randomly cropped into $512\times512$ pixels. The batch size is set to $8$, and the number of the training step is $40$K. We set the temperature parameter $\mathcal{T}=4$, the loss weight $\alpha=3$ for the logits map, and $\alpha=50$ for the feature map for all experiments. For object detection, we employ the same teacher and student networks and the training settings as in~\cite{Zhang2021iod}.

\subsection{Comparison  with %
Recent
Knowledge Distillation Methods}
To verify the effectiveness of our proposed channel-wise
distillation, we compare our method with current distillation methods listed below:

\begin{itemize}
\itemsep -0.12cm

\item
Attention Transfer (AT)~\cite{sergey17paying}: Sergey \etal calculate the summation of all channels %
at
each spatial location to obtain a single channel attention map. $L_2$ is employed to minimize the difference between the attention map.

\item
Local affinity (LOCAL)~\cite{xie18improving}: For each pixel, a local similarity map is constructed, which considers the correlations between itself and its 8 neighborhood pixels. $L_2$ is employed to minimize the difference between the local affinity map.

\item
Pixel-wise distillation (PI)~\cite{liu19structuredDense,liu2019structuredseg,wang20intraclass,chen20fasterseg}: KL divergence is used to align the distribution of each spatial location from two networks.

\item
Pair-wise distillation (PA)~\cite{liu19structuredDense,he19knowledge,liu2019structuredseg}: The correlations between all pixel pairs are considered.
\item
Intra-class feature variation distillation (IFVD)~\cite{wang20intraclass}: The set of similarity between the feature of each pixel and its corresponding class-wise prototype is regarded as the intra-class feature variation to transfer the structural knowledge.

\item
Holistic distillation (HO) \cite{liu19structuredDense,liu2019structuredseg,%
wang20intraclass}: The holistic embeddings of feature maps are computed by a discriminator, which is used to minimize the discrepancy between high-order relations.

\end{itemize}
\begin{table*}[htb]%
	\renewcommand\arraystretch{1.1}
	\footnotesize
	\centering

	\begin{tabular}{c|c|c|c|c|c|c|c|c|c|c}
   \hline
   \textcolor{white}{*}Method\textcolor{white}{*} &\textcolor{white}{*}mIoU\textcolor{white}{*}        &\textcolor{white}{**}road\textcolor{white}{**}       &\textcolor{white}{*}sidewalk\textcolor{white}{*}      &\textcolor{white}{*}building\textcolor{white}{*}      &\textcolor{white}{**}wall\textcolor{white}{**}       &\textcolor{white}{**}fence\textcolor{white}{**}      &\textcolor{white}{**}pole\textcolor{white}{**}       &traffic light      &traffic sign      &vegetation \\
   \hline
   PA    &71.41       &97.30      &80.48         &90.76         &37.89      &52.78      &60.33      &63.48              &74.06             &91.69 \\
   IFVD  &71.66       &97.56      &81.44         &{\bf91.49}    &44.45      &{\bf55.95} &62.40      &66.38              &76.44             &91.85 \\
  \hline
   \textbf{Ours}    &{\bf75.13}  &{\bf97.64} &{\bf81.97}    &91.89         &{\bf49.44} &56.84      &{\bf62.53} &{\bf68.73}         &{\bf77.60}        &{\bf92.20} \\
  \hline
   Class &terrain      &sky        &person      &rider      &car        &truck      &bus        &train      &motorcycle      &bicycle \\
\hline
   PA    &58.60        &93.48      &78.96       &55.45      &93.42      &63.79      &78.48      &60.12      &{\bf51.62}      &74.01 \\
   IFVD  &61.29        &93.97      &78.64       &52.33      &93.50      &60.25      &74.70      &58.81      &44.85           &75.41 \\
\hline
   \textbf{Ours}   &{\bf63.37}   &{\bf94.32} &{\bf80.06}  &{\bf58.49} &{\bf94.18} &{\bf70.31} &{\bf85.61} &{\bf72.85} &52.92           &{\bf76.58} \\
\hline
	\end{tabular}
	   \caption{The class IoU of our proposed channel-wise
	   distillation method compared with %
	   other two typical structural knowledge transfer methods on the validation set of Cityscape, where PSPNet-R18 (1.0)
	   is used
	   as the student network.
	   }
	\label{tab:class-iou-table}
\end{table*}
We apply all these popular distillation methods to both the inner feature map and the final logits map. The conventional cross-entropy loss is applied in all experiments. The computational complexity and performance of spatial distillation methods are reported in Table~\ref{tab:individual-competition-table}.

Given the input feature map (logits map)
of
the size of $h_{f}\times w_{f}\times c$ ($h_{s}\times w_{s}\times n$), where $h_{f} (h_{s}) \times w_{f} (w_{s})$ is the shape of the feature map (logits map). $c$ is the number of channels and $n$ is the number of classes.

As reported  in Table \ref{tab:individual-competition-table}, all distillation methods can improve the performance of the student network. Our channel %
distillation method outperforms all spatial distillation methods.
Ours outperforms the best spatial distillation method (AT) by $2.5\%$. Moreover, %
our method
is more efficient as it requires less computational cost than other methods during the training phase.

Furthermore, we list the detailed class IoU of our method and two recent state-of-the-art methods, PA~\cite{liu19structuredDense} and IFVD~\cite{hou20inter} in Table \ref{tab:class-iou-table}.  These methods propose to transfer structure information in semantic segmentation. Our methods significantly improve the class accuracy of several objects, such as traffic light, terrain, wall, truck, bus, and train, indicating that the channel distribution can well
transfer the structural knowledge.

\subsection{Ablation Study}

We show the effectiveness of the channel-wise
distillation and discuss the choice of the hyper-parameters in semantic segmentation in this section. The baseline student model is PSPNet-R18, and the teacher model is the PSPNet-R101. All the results are evaluated on the validation set of Cityscapes.

\noindent\textbf{Effectiveness of channel-wise
distillation.} The normalized channel-wise
probability map
and the asymmetric KL divergence play an important role in our distillation method. We conduct experiments with four different variants to show the effectiveness of proposed methods in Table~\ref{tab:abl}.

All the distillation methods are applied to the same activation maps as input; and we %
use
the same training scheme as described in Section~\ref{sec:imp}.

`PI' represents the pixel-level knowledge distillation, which normalizes the activation of each spatial location.
`$L_2$ w/o NORM' represents that we directly minimize the difference between the feature maps from two networks, which considers the difference at all locations in all channels equally. `Bhat' is
the
Bhattacharyya distance~\cite{bhattacharyya1943measure}, which is a symmetrical distribution measurement.
It aligns the discrepancy in each channel.

From Table~\ref{tab:abl}, we can see that the asymmetric KL divergence %
measuring the normalized channel %
discrepancy
achieves the best performance. Note that as the KL divergence is asymmetric, the input of the student and teacher can not be swapped.
We experiment by changing the order of the input in the KL divergence, and the training does not converge.

\begin{table}[b]
\centering
\footnotesize
\begin{tabular}{l|c|c|c|c}
\hline
Method   & Norm.\  & Divergence &Logits map& Feature map\\
\hline
Teacher  & -         & -          &   78.56   & 78.56\\
Student  & -         & -          &   69.10   &69.10\\
\hline
PI       & Spatial   & KL         &    71.74  & 70.02\\
$L_2$ w/o norm.\    & None      & MSE  &70.83   &71.37      \\
$L_2$    & Channel   & MSE        & 71.60  & 71.57   \\
Bhat     & Channel   & Bhat       & 72.21  & 71.96    \\
Ours     & Channel   & KL         & \textbf{74.87}& \textbf{74.27} \\
\hline
\end{tabular}
\caption{Mean IoU on the Cityscapes validation set. We can see that with the channel normalization and the asymmetry KL divergence, the proposed channel-wise
distillation achieves the best performance among other variants. All the results are the %
average
of three runs.}
\label{tab:abl}
\end{table}

\begin{figure}[t]
  \centering
      \includegraphics[width=.81\linewidth]{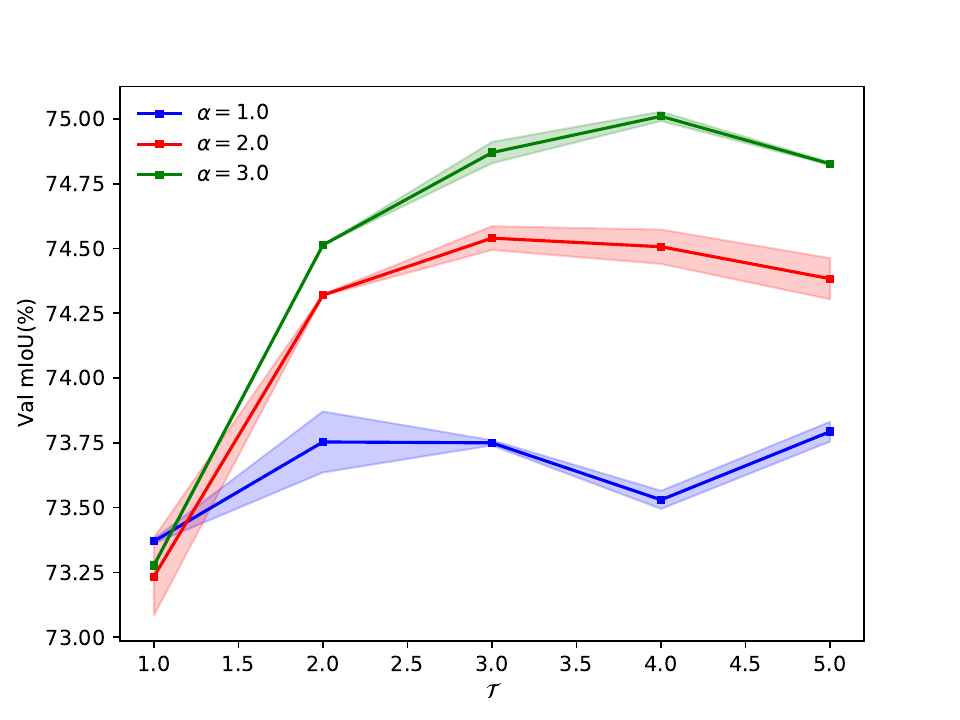}
   \caption{Impact of the temperature parameter $\mathcal{T}$ and the loss weight %
   $\alpha $.
   }
   \label{fig:abl}
\end{figure}

\noindent\textbf{Impact of the temperature parameter and loss weights.} We conduct experiments to %
vary
the channel-wise %
probability maps
by adjusting the temperature parameter $\mathcal{T}$ under different loss weights $\alpha$. The experiments are conducted on the logits map. Results are illustrated in Figure~\ref{fig:abl}.

All the results are the mean of three runs. The loss weight is set to $1,2,3$, and $\mathcal{T} \in [1,5]$. The distribution tends to be softer if we increase $\mathcal{T}$.

From the figure, we can see that a softer
probability map
may help the knowledge distillation. Besides, in a certain range, the performance is stable. The performance %
appears to
drop %
if $\mathcal{T}$ is %
set to be
small. In such cases, %
the method
only focuses on limited salient pixels. We %
attain
the best performance when $\mathcal{T}=4$ and $\alpha=3$ with the PSPNet18 on the Cityscapes validation set.

\begin{figure}[!t]
\centering
		\includegraphics[width=1.0\linewidth]{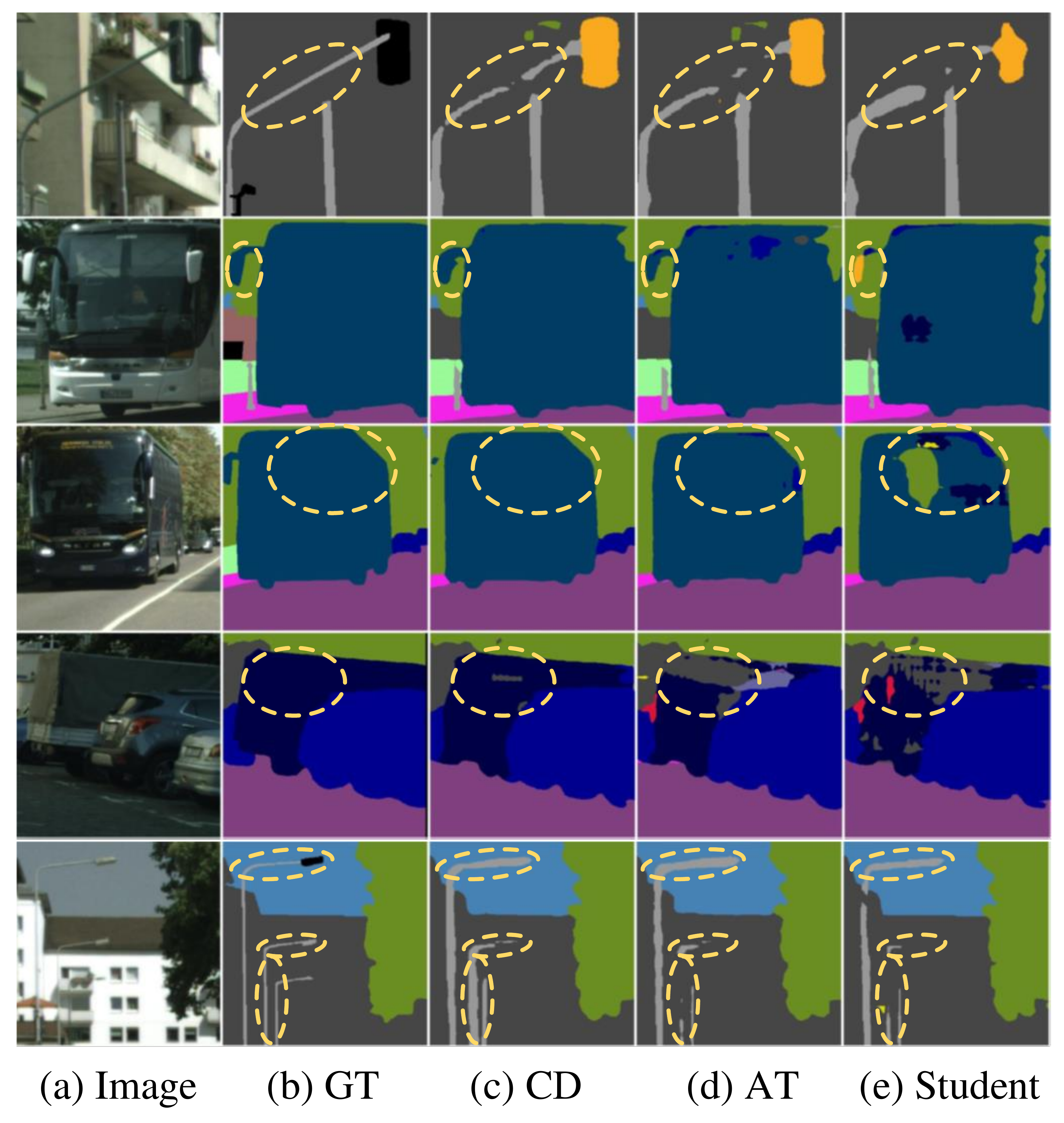}
	\caption{\textbf{Qualitative segmentation results } on Cityscapes %
	of the
	PSPNet-R18 model: (a) raw images, (b) ground truth (GT), (c) channel-wise
	distillation (CD), (d) the best spatial distillation schemes: attention transfer (AT); and (e) the output of the original student model without KD. }
	\label{fig:compare}
\end{figure}

   \begin{table}[htb]
      \renewcommand\arraystretch{1.1}

      \footnotesize

      \centering

      \begin{tabular}{p{2.2cm}|p{1.0cm}|p{1.0cm}|p{0.6cm}|p{0.6cm}}
         \hline

         \multirow{2}{*}{Method} &\multicolumn{1}{c|}{\multirow{2}{*}{Params (M)}} &\multicolumn{1}{c|}{\multirow{2}{*}{FLOPs (G)}} &\multicolumn{2}{c}{mIoU (\%)} \\
         \cline{4-5}
         & & &\multicolumn{1}{c|}{Val} &\multicolumn{1}{c}{Test}              \\
   \hline
         \multicolumn{1}{l|}{ENet \cite{adam16enet}}          &\multicolumn{1}{c|}{0.358} &\multicolumn{1}{c|}{3.612} &\multicolumn{1}{c|}{$-$} &\multicolumn{1}{c}{$58.3$}   \\
         \multicolumn{1}{l|}{ESPNet \cite{sachin18espnet}}    &\multicolumn{1}{c|}{0.363} &\multicolumn{1}{c|}{4.422} &\multicolumn{1}{c|}{$-$} &\multicolumn{1}{c}{$60.3$}   \\
         \multicolumn{1}{l|}{ERFNet \cite{eduardo17erfnet}}   &\multicolumn{1}{c|}{2.067} &\multicolumn{1}{c|}{25.60} &\multicolumn{1}{c|}{$-$} &\multicolumn{1}{c}{$68.0$}   \\
         \multicolumn{1}{l|}{ICNet \cite{zhao18icnet}}        &\multicolumn{1}{c|}{26.50} &\multicolumn{1}{c|}{28.30} &\multicolumn{1}{c|}{$-$} &\multicolumn{1}{c}{$69.5$}   \\
         \multicolumn{1}{l|}{FCN \cite{jonathan15fcn}}        &\multicolumn{1}{c|}{134.5} &\multicolumn{1}{c|}{333.9} &\multicolumn{1}{c|}{$-$} &\multicolumn{1}{c}{$62.7$} \\
         \multicolumn{1}{l|}{RefineNet \cite{Lin17RefineNet}} &\multicolumn{1}{c|}{118.1} &\multicolumn{1}{c|}{525.7} &\multicolumn{1}{c|}{$-$} &\multicolumn{1}{c}{$73.6$}   \\
         \multicolumn{1}{l|}{OCNet \cite{yuan2018ocnet}}        &\multicolumn{1}{c|}{62.58} &\multicolumn{1}{c|}{548.5} &\multicolumn{1}{c|}{$-$} &\multicolumn{1}{c}{$80.1$}   \\
   \hline
         \multicolumn{5}{c}{Results w/ and w/o distillation schemes}\\
    \hline

         \multicolumn{1}{l|}{T:PSPNet \cite{zhao17pspnet}} &\multicolumn{1}{c|}{70.43} &\multicolumn{1}{c|}{574.9} &\multicolumn{1}{c|}{$78.5$} &\multicolumn{1}{c}{$78.4$}   \\
  \hline
         \multicolumn{1}{l|}{S:PSPNet-R18$^{\diamond}$(0.5)}      &\multicolumn{1}{c|}{$3.271$} &\multicolumn{1}{c|}{31.53} &\multicolumn{1}{c|}{$61.17$} &\multicolumn{1}{c}{$-$}   \\
         \multicolumn{1}{l|}{+SKDS \cite{liu2019structuredseg}} &\multicolumn{1}{c|}{$3.271$} &\multicolumn{1}{c|}{31.53} &\multicolumn{1}{c|}{$61.60$} &\multicolumn{1}{c}{$60.50$}   \\
         \multicolumn{1}{l|}{+SKDD \cite{liu19structuredDense}} &\multicolumn{1}{c|}{$3.271$} &\multicolumn{1}{c|}{31.53} &\multicolumn{1}{c|}{$62.35$} &\multicolumn{1}{c}{$-$}   \\
         \multicolumn{1}{l|}{+IFVD \cite{wang20intraclass}}     &\multicolumn{1}{c|}{$3.271$} &\multicolumn{1}{c|}{31.53} &\multicolumn{1}{c|}{$63.35$} &\multicolumn{1}{c}{$63.68$}   \\
         \multicolumn{1}{l|}{+Ours-feaure}                             &\multicolumn{1}{c|}{$3.271$} &\multicolumn{1}{c|}{31.53} &\multicolumn{1}{c|}{$63.06$} &\multicolumn{1}{c}{$63.12$}   \\
         \multicolumn{1}{l|}{+Ours-logits}                             &\multicolumn{1}{c|}{$3.271$} &\multicolumn{1}{c|}{31.53} &\multicolumn{1}{c|}{$68.57$} &\multicolumn{1}{c}{$66.75$}   \\

 \hline
         \multicolumn{1}{l|}{S:PSPNet-R18$^{\diamond}$}      &\multicolumn{1}{c|}{$13.07$} &\multicolumn{1}{c|}{125.8} &\multicolumn{1}{c|}{$63.63$} &\multicolumn{1}{c}{$-$}   \\
         \multicolumn{1}{l|}{+SKDS \cite{liu2019structuredseg}} &\multicolumn{1}{c|}{$13.07$} &\multicolumn{1}{c|}{125.8} &\multicolumn{1}{c|}{$63.20$} &\multicolumn{1}{c}{$62.10$}   \\
         \multicolumn{1}{l|}{+SKDD \cite{liu19structuredDense}} &\multicolumn{1}{c|}{$13.07$} &\multicolumn{1}{c|}{125.8} &\multicolumn{1}{c|}{$64.68$} &\multicolumn{1}{c}{$-$}   \\
         \multicolumn{1}{l|}{+IFVD \cite{wang20intraclass}}     &\multicolumn{1}{c|}{$13.07$} &\multicolumn{1}{c|}{125.8} &\multicolumn{1}{c|}{$66.63$} &\multicolumn{1}{c}{$65.72$}   \\
         \multicolumn{1}{l|}{Ours-feature}                              &\multicolumn{1}{c|}{$13.07$} &\multicolumn{1}{c|}{125.8} &\multicolumn{1}{c|}{$66.85$} &\multicolumn{1}{c}{$66.03$}   \\
        \multicolumn{1}{l|}{Ours-logits}                              &\multicolumn{1}{c|}{$13.07$} &\multicolumn{1}{c|}{125.8} &\multicolumn{1}{c|}{$71.03$} &\multicolumn{1}{c}{$70.43$}   \\
\hline
         \multicolumn{1}{l|}{S:PSPNet-R18$^{\star}$}         &\multicolumn{1}{c|}{$13.07$} &\multicolumn{1}{c|}{125.8} &\multicolumn{1}{c|}{$70.09$} &\multicolumn{1}{c}{$67.60$}   \\
         \multicolumn{1}{l|}{+SKDS \cite{liu2019structuredseg}} &\multicolumn{1}{c|}{$13.07$} &\multicolumn{1}{c|}{125.8} &\multicolumn{1}{c|}{$72.70$} &\multicolumn{1}{c}{$71.40$}   \\
         \multicolumn{1}{l|}{+SKDD \cite{liu19structuredDense}} &\multicolumn{1}{c|}{$13.07$} &\multicolumn{1}{c|}{125.8} &\multicolumn{1}{c|}{$74.08$} &\multicolumn{1}{c}{$-$}   \\
         \multicolumn{1}{l|}{+IFVD \cite{wang20intraclass}}     &\multicolumn{1}{c|}{$13.07$} &\multicolumn{1}{c|}{125.8} &\multicolumn{1}{c|}{$74.54$} &\multicolumn{1}{c}{$72.74$}   \\
         \multicolumn{1}{l|}{+Ours-feature}                             &\multicolumn{1}{c|}{$13.07$} &\multicolumn{1}{c|}{125.8} &\multicolumn{1}{c|}{$74.63$} &\multicolumn{1}{c}{$73.22$}   \\
        \multicolumn{1}{l|}{+Ours-logits}                             &\multicolumn{1}{c|}{$13.07$} &\multicolumn{1}{c|}{125.8} &\multicolumn{1}{c|}{$75.90$} &\multicolumn{1}{c}{$74.58$}   \\

\hline
\hline
         \multicolumn{1}{l|}{S:Deeplab-R18$^{\diamond}$(0.5)}  &\multicolumn{1}{c|}{$3.15$} &\multicolumn{1}{c|}{31.06}  &\multicolumn{1}{c|}{$61.83$} &\multicolumn{1}{c}{$60.51$}   \\
         \multicolumn{1}{l|}{+SKDS \cite{liu2019structuredseg}} &\multicolumn{1}{c|}{$3.15$} &\multicolumn{1}{c|}{31.06}  &\multicolumn{1}{c|}{$62.71$} &\multicolumn{1}{c}{$61.69$}   \\
         \multicolumn{1}{l|}{+IFVD \cite{wang20intraclass}}     &\multicolumn{1}{c|}{$3.15$} &\multicolumn{1}{c|}{31.06}  &\multicolumn{1}{c|}{$63.12$} &\multicolumn{1}{c}{$62.37$}   \\
         \multicolumn{1}{l|}{+Ours-feature}                             &\multicolumn{1}{c|}{$3.15$} &\multicolumn{1}{c|}{31.06}  &\multicolumn{1}{c|}{$64.61$} &\multicolumn{1}{c}{$63.18$}   \\
        \multicolumn{1}{l|}{+Ours-logits}                             &\multicolumn{1}{c|}{$3.15$} &\multicolumn{1}{c|}{31.06}  &\multicolumn{1}{c|}{$67.44$} &\multicolumn{1}{c}{$67.12$}   \\
\hline
         \multicolumn{1}{l|}{S:Deeplab-R18$^{\star}$}     &\multicolumn{1}{c|}{$12.62$} &\multicolumn{1}{c|}{123.9} &\multicolumn{1}{c|}{$73.37$} &\multicolumn{1}{c}{$72.39$}   \\
         \multicolumn{1}{l|}{+SKDS \cite{liu2019structuredseg}} &\multicolumn{1}{c|}{$12.62$} &\multicolumn{1}{c|}{123.9} &\multicolumn{1}{c|}{$73.87$} &\multicolumn{1}{c}{$72.63$}   \\
         \multicolumn{1}{l|}{+IFVD \cite{wang20intraclass}}     &\multicolumn{1}{c|}{$12.62$} &\multicolumn{1}{c|}{123.9} &\multicolumn{1}{c|}{$74.09$} &\multicolumn{1}{c}{$72.97$}   \\
         \multicolumn{1}{l|}{+Ours-feature}                             &\multicolumn{1}{c|}{$12.62$} &\multicolumn{1}{c|}{123.9} &\multicolumn{1}{c|}{$74.24$} &\multicolumn{1}{c}{$72.56$}   \\
        \multicolumn{1}{l|}{+Ours-logits}                             &\multicolumn{1}{c|}{$12.62$} &\multicolumn{1}{c|}{123.9} &\multicolumn{1}{c|}{$75.91$} &\multicolumn{1}{c}{$74.32$}   \\
\hline

      \end{tabular}
            \caption{Comparison of student variants with the state-of-the-art distillation methods on the Cityscapes dataset.
            $^{\diamond}$ %
            means
            that
            the models are
            trained from scratch, and $^{\star}$ indicates %
            that models are
            initialized by the weights pre-trained on ImageNet.
            R18 %
            stands for
            Resnet18.}
      \label{tab:result-cityscape}
   \end{table}
   \begin{table*}[htb]
  \renewcommand\arraystretch{1.1}
  \footnotesize
  \centering
  \begin{tabular}{c|c|c|c|c|c|c|c|c|c|c}
     \toprule
     \multicolumn{2}{c|}{Model} &Backbone &AP (\%) &AP$_{50}$ &AP$_{75}$ &AP$_{S}$ &AP$_{M}$ &AP$_{L}$ &FPS &Params.\  \\
     \midrule
      \multirow{6}{*}{Two-stage %
      detector
      } &{Faster RCNN} &\multirow{6}{*}{R50} &38.4 &59.0 &42.0 &21.5 &42.1 &50.3 &18.1 &43.57 \\
      &\multicolumn{1}{l|}{+Chen et al.  \cite{Chen2017leod}} & &38.7 &59.0 &42.1 &22.0 &41.9 &51.0 &18.1 &43.57 \\
      &\multicolumn{1}{l|}{+Wang et al.  \cite{Wang2019dod}} & &39.1 &59.8 &42.8 &22.2 &42.9 &51.1 &18.1 &43.57 \\
      &\multicolumn{1}{l|}{+Heo et al.  \cite{Heo2019aco}}  & &38.9 &60.1 &42.6 &21.8 &42.7 &50.7 &18.1 &43.57 \\
      &\multicolumn{1}{l|}{+Zhang et al. \cite{Zhang2021iod}}                   & &41.5 &62.2 &45.1 &23.5 &45.0 &55.3 &18.1 &43.57 \\
      &\multicolumn{1}{l|}{+Our Method}            & &41.7 &62.0 &45.5 &23.3 &45.5 &55.5 &18.1 &43.57 \\
     \midrule
      \multirow{4}{*}{One-stage %
      detector
      } &{RetinaNet} &\multirow{4}{*}{R50} &37.4 &56.7 &39.6 &20.0 &40.7 &49.7 &20.0 &36.19 \\
      &\multicolumn{1}{l|}{+Heo et al.\  \cite{Heo2019aco}} & &37.8 &58.3 &41.1 &21.6 &41.2 &48.3 &20.0 &36.19 \\
      &\multicolumn{1}{l|}{+Zhang et al.\ \cite{Zhang2021iod}}                  & &39.6 &58.8 &42.1 &22.7 &43.3 &52.5 &20.0 &36.19 \\
      &\multicolumn{1}{l|}{+Our Method}           & &40.8 &60.4 &43.4 &22.7 &44.5 &55.3 &20.0 &36.19 \\
     \midrule
      \multirow{3}{*}{Anchor-free detector} &{RepPoints} &\multirow{3}{*}{R50} &38.6 &59.6 &41.6 &22.5 &42.2 &50.4 &18.2 &36.62 \\
      &\multicolumn{1}{l|}{+Zhang et al.\  \cite{Zhang2021iod}}                & &40.6 &61.7 &43.8 &23.4 &44.6 &53.0 &18.2 &36.62 \\
      &\multicolumn{1}{l|}{+Our Method}         & &42.0 &63.0 &45.3 &24.1 &46.1 &55.0 &18.2 &36.62 \\
     \bottomrule
  \end{tabular}
  \caption{Comparison between our methods and other distillation methods on object detection.}
  \label{tab:detec}
\end{table*}

\subsection{Semantic Segmentation%
}

We demonstrate that our proposed channel wise distillation method can be combined with previous semantic segmentation distillation methods, {\it i.e.}, structural knowledge distillation for segmentation/dense prediction (SKDS \cite{liu2019structuredseg} and SKDD \cite{liu19structuredDense}) and intra-class feature variation distillation (IFVD \cite{wang20intraclass}), under various student networks.

We use the proposed channel-wise
distillation both on the logits map (Ours-logits) and the feature map (Ours-feature). The pixel-wise distillation (PI) and the holistic distillation (HO) on the logits map are also included following previous methods ~\cite{liu19structuredDense,hou20inter}.

We first evaluate the performance of our method on the Cityscapes dataset. Various student networks with different encoders and decoders are used to verify the effectiveness of our method. Encoders include ResNet18 (initialized with or without the weights pre-trained on ImageNet, and a channel-halved variant of ResNet18~\cite{he16resnet}), and decoders include PSPhead~\cite{zhao17pspnet} and ASPPhead~\cite{Chen17deeplabv3}. Table \ref{tab:result-cityscape} shows the results on Cityscapes. Experiment results on Pascal VOC~\cite{mark10voc} and ADE20K~\cite{zhou17ade20k} are shown in the supplementary materials.

Our method outperforms SKD and IFVD on five student networks and three benchmarks, which further indicates that the channel-wise
distillation is effective for semantic segmentation.

For the student with the same architectural type as the teacher, {\it i.e.}, PSPNet-R18$^{\diamond}$(0.5), PSPNet-R18$^{\diamond}$ and PSPNet-R18$^{\star}$, the improvements are more significant. As for the student with different architectural types with the teacher, {\it i.e.}, Deeplab-R18$^{\diamond}$(0.5) and  Deeplab-R18$^{\star}$, our method achieves consistent improvement compared with SKDS and IFVD.
Thus, our method works
well with different teacher and student networks.

The student network of %
a
compact model capacity
(PSPNet-R18$^{\diamond}$(0.5)) %
shows
inferior distillation performance (68.57\%) compared to the student with  a larger  %
capacity
(PSPNet-R18$^{\star}$) (75.90\%). This may be attributed to the fact that the capability of small networks is limited compared with the teacher network and can not sufficiently absorb the knowledge of the current task.
For PSPNet-R18, the student initialized by the weights trained on ImageNet obtains the best distillation performance (improved from 70.09\% to 75.90\%), further demonstrating that the well-initialized parameters %
help
the distillation. Thus, the better student
lead to better distillation performance, but the improvement is
less significant
as the gap between the teacher and student network is smaller.

\subsection{Object Detection}

We also
apply
our channel-wise  %
distillation method on the object detection task. The experiments are conducted on MS COCO2017~\cite{lin2014microsoft}.

Various student networks under different paradigms, {\it i.e.}, a two-stage anchor-based method (Faster RCNN~\cite{ren2016faster}), a one-stage anchor-based method (RetinaNet~\cite{lin2017retinanet}) and anchor-free method (RepPoints~\cite{yang2019reppoints}), are used to validate the effectiveness of our method. To make a fair comparison, we experiment on the same teacher with the same hyper-parameters as in \cite{Zhang2021iod}. %

The only modification is that
the feature alignment is changed to our channel-wise
distillation. The results are %
reported
in Table~\ref{tab:detec}. From the table, we can see that our methods achieve consistent improvements (about $3.4\%$ mAP) on strong baseline student networks.
Compared with previous state-of-the-art distillation methods~\cite{Zhang2021iod}, our simple channel-wise
distillation performs better,  especially with anchor-free methods. We improve the RepPoint by $3.4\%$ while Zhang \etal improve the RepPoint by $2\%$. Besides, we can see that the proposed  distillation method can improve $AP_{75}$ more significantly.

\section{Conclusion}
In this paper, we have proposed a novel channel-wise
distillation for dense prediction tasks. Different from previous spatial distillation methods, we normalize the activations %
of each channel to %
a probability map.
Then, the asymmetry KL divergence is applied to minimize  the %
discrepancy
between the teacher and the student network. Experimental results show that the proposed %
distillation method consistently outperforms state-of-the-art distillation methods on four public benchmark datasets with various network backbones, for both  semantic segmentation and object detection.

Additionally, %
our ablation
experiments
demonstrate the efficiency and effectiveness of our channel-wise  %
distillation, and it can further complement the spatial distillation methods. We hope that the proposed simple and effective
distillation method  can serve as a %
strong
baseline for %
effectively training
compact networks
for many other dense prediction tasks, including instance segmentation, depth estimation and panoptic segmentation.









\appendix

\section*{Appendix}
\begin{figure}[htp]
\centering 
		\includegraphics[width=1.0\linewidth]{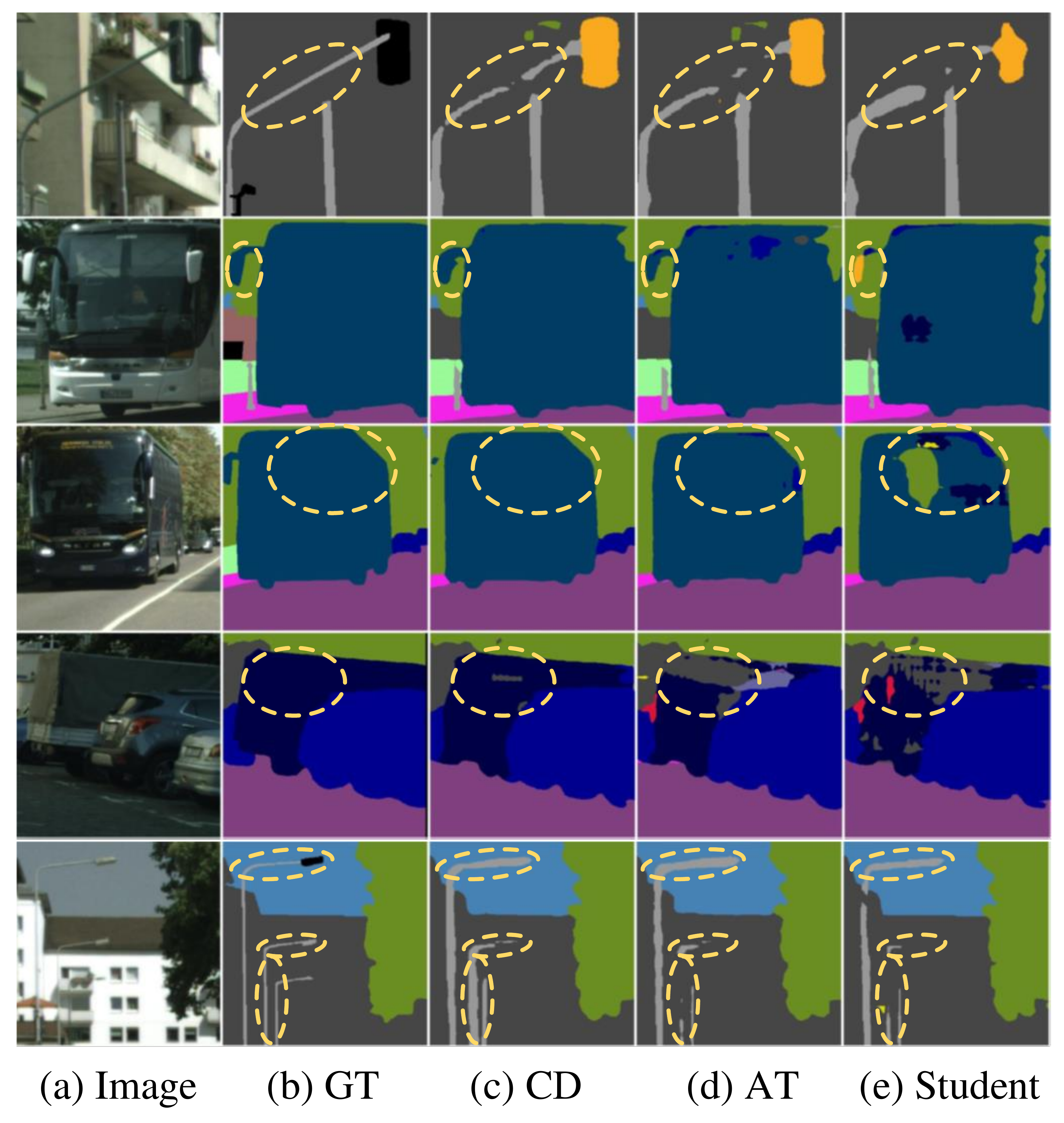}
	\caption{\textbf{Qualitative segmentation results } on Cityscapes 
	of the 
	PSPNet-R18 model: (a) raw images, (b) ground truth (GT), (c) channel-wise  
	distillation (CD), (d) the best spatial distillation schemes: attention transfer (AT); and (e) the output of the original student model without KD. }
	\label{fig:compare}
\end{figure}

\section{Results on Pascal VOC and ADE20K}
To further demonstrate the effectiveness of the proposed channel distribution distillation, we only employ the proposed CD on the feature maps as our final results on  Pascal VOC and ADE20K. The experiment results  are reported in Table~\ref{tab:result-voc} and Table~\ref{tab:result-ade20k}. Multi student-network variants with different encoders and decoders are used to validate the efficiency of our method. Here, encoders include ResNet18 and MobileNetV2, and decoders include PSP-head and ASPP-head.

\begin{table}[!htb]
  \renewcommand\arraystretch{1.1}
  \footnotesize
  \centering
  \begin{tabular}{p{2.4cm}|p{1.0cm}|p{1.5cm}|p{1.5cm}}
     \toprule
     \multirow{1}{*}{Method} &\multirow{1}{*}{Params} &\textcolor{white}{**}{mIoU(\%)} &\textcolor{white}{**}{mAcc(\%)} \\
     \midrule
     \multicolumn{1}{l|}{FCN \cite{jonathan15fcn}}           &\multicolumn{1}{c|}{134.5}       &\multicolumn{1}{c|}{$69.9$}   &\multicolumn{1}{c}{$78.1$}   \\
     \multicolumn{1}{l|}{DeepLabV3 \cite{Chen17deeplabv3}}   &\multicolumn{1}{c|}{87.1}        &\multicolumn{1}{c|}{$77.9$}   &\multicolumn{1}{c}{$85.7$}      \\
     \multicolumn{1}{l|}{PSANet \cite{zhao2018psanet}}       &\multicolumn{1}{c|}{78.13}       &\multicolumn{1}{c|}{$77.9$}   &\multicolumn{1}{c}{$86.6$}      \\
     \multicolumn{1}{l|}{GCNet \cite{cao2019gcnet}}          &\multicolumn{1}{c|}{68.82}       &\multicolumn{1}{c|}{$77.8$}   &\multicolumn{1}{c}{$85.9$}      \\
     \multicolumn{1}{l|}{ANN \cite{zhu19ann}}                &\multicolumn{1}{c|}{65.2}        &\multicolumn{1}{c|}{$76.7$}   &\multicolumn{1}{c}{$84.5$}      \\
     \multicolumn{1}{l|}{OCRNet \cite{YuanCW20}}             &\multicolumn{1}{c|}{70.37}        &\multicolumn{1}{c|}{$80.3$}   &\multicolumn{1}{c}{$87.1$}      \\
     \midrule
     \multicolumn{4}{c}{Results w/ and w/o our distillation schemes}\\
     \midrule
     \multicolumn{1}{l|}{T:PSPNet \cite{zhao17pspnet}}       &\multicolumn{1}{c|}{$70.43$}     &\multicolumn{1}{c|}{$78.52$}    &\multicolumn{1}{c}{$79.57$}      \\
     \midrule
     \multicolumn{1}{l|}{S:PSPNet-R18}                       &\multicolumn{1}{c|}{$13.07$}     &\multicolumn{1}{c|}{$65.42$}    &\multicolumn{1}{c}{$80.43$}       \\
     \multicolumn{1}{l|}{+SKDS \cite{liu2019structuredseg}}  &\multicolumn{1}{c|}{$13.07$}     &\multicolumn{1}{c|}{$67.73$}    &\multicolumn{1}{c}{$81.73$}       \\
     \multicolumn{1}{l|}{+IFDV \cite{wang20intraclass}}      &\multicolumn{1}{c|}{$13.07$}     &\multicolumn{1}{c|}{$68.04$}    &\multicolumn{1}{c}{$82.25$}       \\
     \multicolumn{1}{l|}{+Ours}                              &\multicolumn{1}{c|}{$13.07$}     &\multicolumn{1}{c|}{$69.25$}    &\multicolumn{1}{c}{$83.14$}       \\
     \midrule
     \multicolumn{1}{l|}{S:PSPNet-MBV2}                      &\multicolumn{1}{c|}{$1.98$}      &\multicolumn{1}{c|}{$62.38$}    &\multicolumn{1}{c}{$77.82$}       \\
     \multicolumn{1}{l|}{+SKDS \cite{liu2019structuredseg}}  &\multicolumn{1}{c|}{$1.98$}      &\multicolumn{1}{c|}{$63.95$}    &\multicolumn{1}{c}{$78.93$}       \\
     \multicolumn{1}{l|}{+IFDV \cite{wang20intraclass}}      &\multicolumn{1}{c|}{$1.98$}      &\multicolumn{1}{c|}{$64.73$}    &\multicolumn{1}{c}{$79.81$}       \\
     \multicolumn{1}{l|}{+Ours}                              &\multicolumn{1}{c|}{$1.98$}      &\multicolumn{1}{c|}{$65.93$}    &\multicolumn{1}{c}{$81.45$}       \\
     \midrule
     \multicolumn{1}{l|}{S:Deeplab-R18}                      &\multicolumn{1}{c|}{$12.62$}     &\multicolumn{1}{c|}{$66.81$}    &\multicolumn{1}{c}{$81.14$}       \\
     \multicolumn{1}{l|}{+SKDS \cite{liu2019structuredseg}}  &\multicolumn{1}{c|}{$12.62$}     &\multicolumn{1}{c|}{$68.13$}    &\multicolumn{1}{c}{$82.26$}       \\
     \multicolumn{1}{l|}{+IFDV \cite{wang20intraclass}}      &\multicolumn{1}{c|}{$12.62$}     &\multicolumn{1}{c|}{$68.42$}    &\multicolumn{1}{c}{$82.70$}       \\
     \multicolumn{1}{l|}{+Ours}                              &\multicolumn{1}{c|}{$12.62$}     &\multicolumn{1}{c|}{$69.97$}    &\multicolumn{1}{c}{$83.47$}       \\
     \midrule
     \multicolumn{1}{l|}{S:Deeplab-MBV2}                     &\multicolumn{1}{c|}{$2.45$}      &\multicolumn{1}{c|}{$50.80$}    &\multicolumn{1}{c}{$74.24$}       \\
     \multicolumn{1}{l|}{+SKDS \cite{liu2019structuredseg}}  &\multicolumn{1}{c|}{$2.45$}      &\multicolumn{1}{c|}{$52.11$}    &\multicolumn{1}{c}{$75.17$}       \\
     \multicolumn{1}{l|}{+IFDV \cite{wang20intraclass}}      &\multicolumn{1}{c|}{$2.45$}      &\multicolumn{1}{c|}{$53.39$}    &\multicolumn{1}{c}{$76.02$}       \\
     \multicolumn{1}{l|}{+Ours}                              &\multicolumn{1}{c|}{$2.45$}      &\multicolumn{1}{c|}{$54.62$}    &\multicolumn{1}{c}{$77.13$}       \\
     \bottomrule
  \end{tabular}
  \caption{mIoU and mAcc on validation set of VOC 2012, R18 (MBV2) is the abbreviation for Resnet18 (MobileNetV2).}
  \label{tab:result-voc}
\end{table}
\begin{table}[!htb]
  \renewcommand\arraystretch{1.1}
  \footnotesize

  \centering

  \begin{tabular}{p{2.4cm}|p{1.0cm}|p{1.5cm}|p{1.5cm}}
     \toprule
     \multirow{1}{*}{Method}                                 &\multirow{1}{*}{Params}          &\textcolor{white}{**}{mIoU(\%)} &\textcolor{white}{**}{mAcc(\%)} \\
     \midrule

     \multicolumn{1}{l|}{FCN \cite{jonathan15fcn}}           &\multicolumn{1}{c|}{134.5}       &\multicolumn{1}{c|}{$39.91$}  &\multicolumn{1}{c}{$49.62$} \\
     \multicolumn{1}{l|}{DeepLabV3 \cite{Chen17deeplabv3}}   &\multicolumn{1}{c|}{87.1}        &\multicolumn{1}{c|}{$44.99$}  &\multicolumn{1}{c}{$55.81$}    \\
     \multicolumn{1}{l|}{PSANet \cite{zhao2018psanet}}       &\multicolumn{1}{c|}{78.13}       &\multicolumn{1}{c|}{$43.74$}  &\multicolumn{1}{c}{$54.09$}    \\
     \multicolumn{1}{l|}{GCNet \cite{cao2019gcnet}}          &\multicolumn{1}{c|}{68.82}       &\multicolumn{1}{c|}{$43.68$}  &\multicolumn{1}{c}{$54.28$}    \\
     \multicolumn{1}{l|}{ANN \cite{zhu19ann}}                &\multicolumn{1}{c|}{65.2}        &\multicolumn{1}{c|}{$42.93$}  &\multicolumn{1}{c}{$53.25$}    \\
     \multicolumn{1}{l|}{OCRNet \cite{YuanCW20}}             &\multicolumn{1}{c|}{70.37}        &\multicolumn{1}{c|}{$43.70$}  &\multicolumn{1}{c}{$53.74$}      \\
     \midrule
     \multicolumn{4}{c}{Results w/ and w/o our distillation schemes}\\
     \midrule
     \multicolumn{1}{l|}{T:PSPNet \cite{zhao17pspnet}}       &\multicolumn{1}{c|}{$70.43$}     &\multicolumn{1}{c|}{$44.39$}    &\multicolumn{1}{c}{$45.35$}      \\
     \midrule
     \multicolumn{1}{l|}{S:PSPNet-R18}                       &\multicolumn{1}{c|}{$13.07$}     &\multicolumn{1}{c|}{$24.65$}    &\multicolumn{1}{c}{$33.66$}       \\
     \multicolumn{1}{l|}{+SKDS \cite{liu2019structuredseg}}  &\multicolumn{1}{c|}{$13.07$}     &\multicolumn{1}{c|}{$25.11$}    &\multicolumn{1}{c}{$33.72$}       \\
     \multicolumn{1}{l|}{+IFDV \cite{wang20intraclass}}      &\multicolumn{1}{c|}{$13.07$}     &\multicolumn{1}{c|}{$25.72$}    &\multicolumn{1}{c}{$33.83$}       \\
     \multicolumn{1}{l|}{+Ours}                              &\multicolumn{1}{c|}{$13.07$}     &\multicolumn{1}{c|}{$26.80$}    &\multicolumn{1}{c}{$34.02$}       \\
     \midrule
     \multicolumn{1}{l|}{S:PSPNet-MBV2}                      &\multicolumn{1}{c|}{$1.98$}      &\multicolumn{1}{c|}{$23.15$}    &\multicolumn{1}{c}{$32.93$}       \\
     \multicolumn{1}{l|}{+SKDS \cite{liu2019structuredseg}}  &\multicolumn{1}{c|}{$1.98$}      &\multicolumn{1}{c|}{$24.79$}    &\multicolumn{1}{c}{$34.04$}       \\
     \multicolumn{1}{l|}{+IFDV \cite{wang20intraclass}}      &\multicolumn{1}{c|}{$1.98$}      &\multicolumn{1}{c|}{$25.33$}    &\multicolumn{1}{c}{$35.57$}       \\
     \multicolumn{1}{l|}{+Ours}                              &\multicolumn{1}{c|}{$1.98$}      &\multicolumn{1}{c|}{$27.97$}    &\multicolumn{1}{c}{$37.16$}       \\

     \midrule
     \multicolumn{1}{l|}{S:Deeplab-R18}                      &\multicolumn{1}{c|}{$12.62$}     &\multicolumn{1}{c|}{$24.89$}    &\multicolumn{1}{c}{$33.60$}       \\
     \multicolumn{1}{l|}{+SKDS \cite{liu2019structuredseg}}  &\multicolumn{1}{c|}{$12.62$}     &\multicolumn{1}{c|}{$25.52$}    &\multicolumn{1}{c}{$34.10$}       \\
     \multicolumn{1}{l|}{+IFDV \cite{wang20intraclass}}      &\multicolumn{1}{c|}{$12.62$}     &\multicolumn{1}{c|}{$26.53$}    &\multicolumn{1}{c}{$34.79$}       \\
     \multicolumn{1}{l|}{+Ours}                              &\multicolumn{1}{c|}{$12.62$}     &\multicolumn{1}{c|}{$27.37$}    &\multicolumn{1}{c}{$35.34$}       \\
     \midrule
     \multicolumn{1}{l|}{S:Deeplab-MBV2}                     &\multicolumn{1}{c|}{$2.45$}      &\multicolumn{1}{c|}{$24.98$}    &\multicolumn{1}{c}{$35.34$}       \\
     \multicolumn{1}{l|}{+SKDS \cite{liu2019structuredseg}}  &\multicolumn{1}{c|}{$2.45$}      &\multicolumn{1}{c|}{$26.10$}    &\multicolumn{1}{c}{$36.51$}       \\
     \multicolumn{1}{l|}{+IFDV \cite{wang20intraclass}}      &\multicolumn{1}{c|}{$2.45$}      &\multicolumn{1}{c|}{$27.25$}    &\multicolumn{1}{c}{$37.23$}       \\
     \multicolumn{1}{l|}{+Ours}                              &\multicolumn{1}{c|}{$2.45$}      &\multicolumn{1}{c|}{$29.18$}    &\multicolumn{1}{c}{$38.08$}       \\
     \bottomrule
  \end{tabular}
  \caption{mIoU and mAcc on validation set of ADE20K, R18 (MBV2) is the abbreviation for Resnet18 (MobileNetV2).}
  \label{tab:result-ade20k}
\end{table}

\noindent  {\bf Pascal VOC.} We 
evaluate the performance of our method on the Pascal VOC dataset. The distillation results are listed in Table \ref{tab:result-voc}.
Our proposed CD improves PSPNet-R18 without distillation by 3.83\%, outperforms the SKDS and IFVD by 1.51\% and 1.21\%.
Consistent improvements on other student networks with different encoders and decoders are achieved.
%
The gains on PSPNet-MBV2 with our method is 3.55\%, surpassing the SKDS and IFVD by 1.98\% and 1.20\%.
As for Deeplab-R18, our CD improves the student from 66.81\% to 69.97\%, 
outperforming
the SKDS and IFVD by 1.84\% and 1.55\% 
respectively.
Besides, the performance of Deeplab-MBV2 with our distillation 
is increased
from 50.80\% to 54.62\%, 
outperforming
the SKDS and IFVD by 2.51\% and 1.23\% respectively.

\noindent {\bf ADE20K.} We also evaluate our method on the ADE20K dataset to further demonstrate that CD works better than other structural knowledge distillation methods. The results are 
shown
in Table \ref{tab:result-ade20k}.
Our proposed CD improves PSPNet-R18 without distillation by 3.83\%,
and
outperforms the SKDS and IFVD by 1.51\% and 1.21\% in several.
Notable performance gains on other student with different encoders and decoders are also consistently achieved,
As for PSPNet-MBV2, our method achieves a superior performance of 27.97\%, surpassing the student, SKDS and IFVD by 4.82\%, 3.18\% and 2.64\%.
The gain on Deeplab-R18 with our CD is 2.48\%, 
outperforming
the SKDS and IFVD by 1.85\% and 0.84\%.
Finally, the performance of Deeplab-MBV2 with our channel-wise distillation 
is increased
from 24.98\% to 29.18\%, outperforming  the SKDS and IFVD by 3.08\% and 1.93\% respectively.


\section{More visualization results}

\begin{figure}[!htb]
	\begin{center}
      \includegraphics[width=1.0\linewidth]{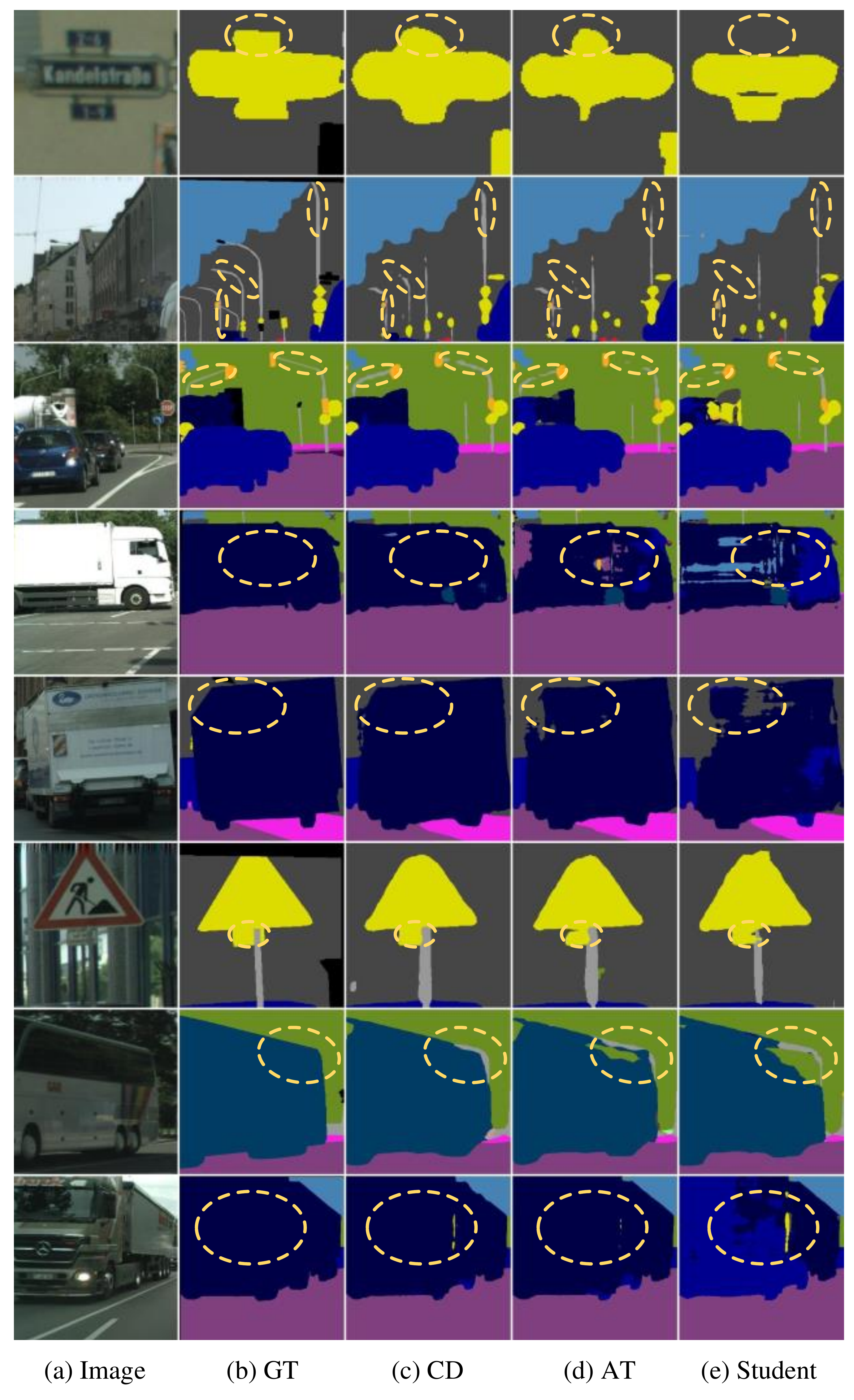}
	\end{center}
	\caption{Qualitative segmentation results on Cityscapes produced from PSPNet-R18: (a) raw images, (b) ground truth (GT), (c) channel-wise distillation (CW), (d) the spatial distillation schemes: attention transfer (AT), and (e) output of the original student model. }
   \label{fig:seg_result}
\end{figure}

\begin{figure}[!htb]
	\begin{center}
      \includegraphics[width=1.0\linewidth]{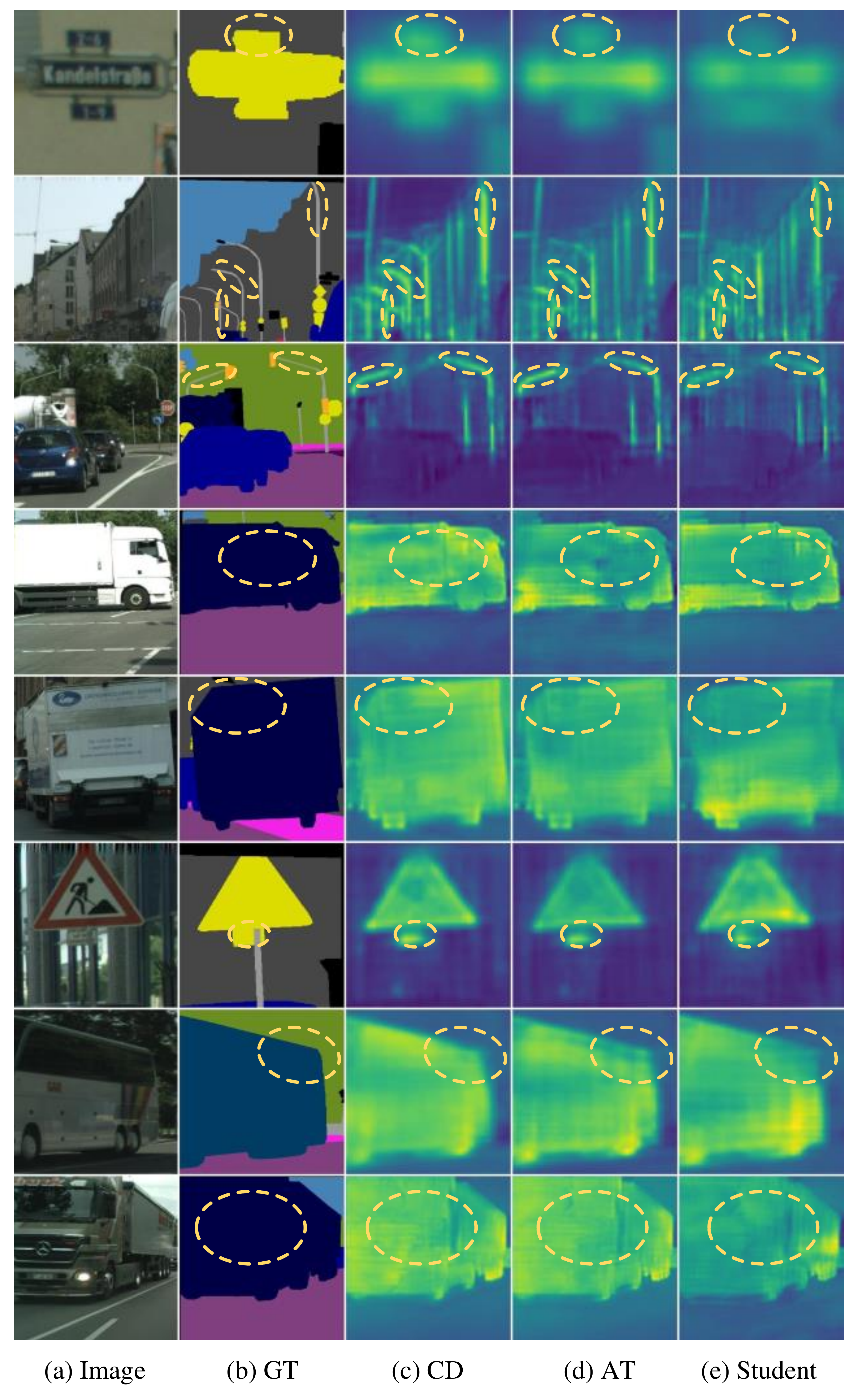}
	\end{center}
   \caption{The channel distribution of the student under three paradigms. (a) raw images, (b) ground truth (GT), (c) channel distillation, (d) the spatial distillation schemes: attention transfer (AT), and (e) output of the original student model.}
   \label{fig:dis_result}
	\vspace{-1em}
\end{figure}

\begin{figure}[!htb]
	\begin{center}
		\includegraphics[width=1.0\linewidth]{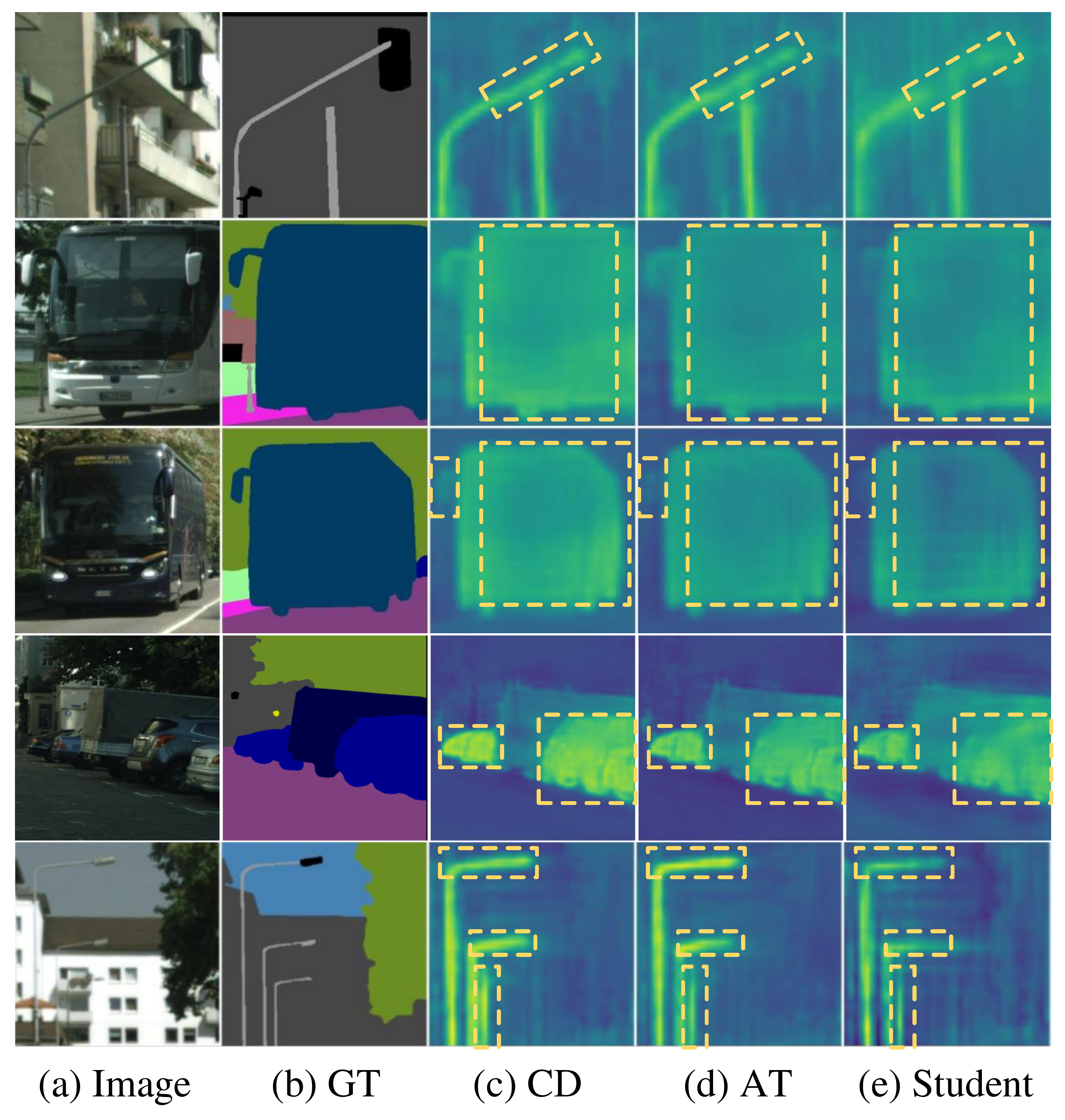}
	\end{center}
	\caption{The channel distribution of the student under three paradigms. The yellow dotted lines show the activation maps of CD are better than that in AT and the student network.}
	\label{fig:channel-wise-fore/back-ground-compare}
	\vspace{-1em}
\end{figure}

We 
list the visualization results in Figure \ref{fig:seg_result} to intuitively demonstrate that,
the channel distribution distillation method (CD) outperforms the spatial distillation strategy (attention transfer).
Besides, to evaluate the effectiveness of the proposed channel distribution distillation,
we visualize the channel distribution of the student network under three paradigms,
{\it i.e.}, original network, distilled by the attention transfer (AT) and channel distribution distillation respectively,
in Figure \ref{fig:dis_result} and Figure~\ref{fig:channel-wise-fore/back-ground-compare}.

{\small
\bibliographystyle{ieee_fullname}
\bibliography{arxiv}
}

\end{document}